%% file: main_arxiv.tex
\documentclass{article}
\usepackage[utf8]{inputenc}

\usepackage[margin=1.3in]{geometry}

\usepackage{times}
\usepackage{amsfonts,psfrag,epsfig}
\usepackage{amsfonts,epsfig}
\usepackage{color}
\usepackage{amsmath}
\usepackage{amssymb}
\usepackage[amsmath,thmmarks]{ntheorem}
\usepackage{tabularx}
\usepackage{algorithm}
\usepackage{algorithmic}
\usepackage{graphicx}
\usepackage{caption}
\usepackage{subcaption}

\newtheorem{theorem}{Theorem}
\newtheorem{lemma}[theorem]{Lemma}

\newtheorem{proposition}[theorem]{Proposition}

\newtheorem{condition}{Condition}

\newtheorem{definition}{Definition}

\newtheorem{assumption}{Assumption}

\theorembodyfont{\normalfont}

\begin{document}

\title{Sequential Local Learning for Latent Graphical Models}

\author{Sejun Park\thanks{S.\ Park and J.\ Shin are
with Department of Electrical Engineering, Korea Advanced Institute of Science \& Technology, Republic of Korea.
Email: sejun.park@kaist.ac.kr, jinwoos@kaist.ac.kr}\and Eunho Yang\thanks{E.\ Yang is with Department of Computer Science, Korea Advanced Institute of Science \& Technology, Republic of Korea.
Email: yangeh@gmail.com}\and Jinwoo Shin$^*$}

\maketitle

\input{abstract_arxiv}
\input{introduction_arxiv}

\input{preliminary_arxiv}
\input{mainresult1_arxiv}

\input{mainresult2_arxiv}

\input{example_arxiv}
\input{conclusion_arxiv}

\bibliography{reference_arxiv}
\bibliographystyle{plain}

\input{appendix_arxiv}


\end{document}

%% file: abstract_arxiv.tex
\begin{abstract} 
Learning parameters of latent graphical models (GM) is inherently much harder than that of 
no-latent ones since the latent variables make the corresponding log-likelihood
non-concave. Nevertheless, expectation-maximization schemes are popularly used in practice, 
but they are typically stuck in local optima. In the recent years, 
the method of moments have provided a refreshing angle for resolving the non-convex issue, but it is applicable
to a quite limited class of latent GMs.
In this paper, we aim for enhancing its power via enlarging such a class of latent GMs. 
To this end, we introduce two novel concepts, coined marginalization and conditioning,
which can reduce the problem of learning a larger GM to that of a smaller one.
More importantly, they lead to a sequential learning framework
that repeatedly increases the learning portion of given latent GM, and
thus covers a significantly broader and more complicated class of loopy latent GMs 
which include convolutional and random regular models. 
\end{abstract} 

%% file: introduction_arxiv.tex
\section{Introduction}
Graphical models (GM) are succinct representation of a joint distribution on a graph where each node corresponds to a random variable and each edge represents the conditional independence between random variables.
GM have been successfully applied for various fields including 
information theory \cite{gallager1962low,kschischang1998iterative},
physics \cite{parisi1988statistical} 
and machine learning \cite{jordan1998learning,freeman2000learning}.
Introducing latent variables to GM
has been popular approaches for enhancing their representation
powers in recent deep models, e.g., 
convolutional/restricted/deep Boltzmann machines \cite{lee2009convolutional,salakhutdinov2009deep}.
Furthermore, they are inevitable in certain 
scenarios when a part of samples is missing, e.g., see \cite{fayyad1996data}. 

However, learning parameters of latent GMs is significantly harder than
that of no-latent ones since 
the latent variables  make  the  corresponding  negative log-likelihood non-convex. 
The main challenge comes from the difficulty of 
inferring unobserved/latent marginal probabilities associated
to latent/hidden variables. 
Nevertheless, the expectation-maximization (EM) schemes \cite{dempster1977maximum}
have been popularly used in practice with empirical successes, 
e.g., contrastive divergence learning for deep models \cite{hinton2002training}.
They iteratively infer unobserved marginals given current estimation
of parameters, and typically stuck at local optima of the log-likelihood function \cite{redner1984mixture}.

To address this issue, 
the spectral methods have provided a refreshing angle
on learning probabilistic latent models \cite{anandkumar2014tensor}.
These theoretical methods exploit the linear algebraic
properties of a model to factorize 
observed (low-order) moments/marginals into unobserved ones.
Furthermore, the factorization methods 
can be combined with convex log-likelihood optimizations
under certain structures, coined {exclusive views},
of latent GMs \cite{chaganty2014estimating}.
Both factorization methods and exclusive views can be understood as 
`local algorithms' handling certain partial structures of latent GMs.
However, up to now,
they are known to be applicable to a quite limited class of latent GMs,
and not as broadly applicable as EM,
which is the main motivation of this paper.



\vspace{0.05in}
\noindent{\bf Contribution.}
Our major question is ``Can we learn latent GMs of more complicated structures beyond naive applications
of local algorithms, e.g., known factorization methods or exclusive views?''.
To address this, we introduce two novel concepts, called marginalization and conditioning,
which reduce the problem of learning a larger GM to that of a smaller one.
Hence, if the smaller one is possible to be processed by 
known local algorithms, 
then the larger one is too.
Our marginalization concept suggests to search a `marginalizable' subset of variables of GM
so that their marginal distributions are invariant with respect to other variables
under certain graphical transformations.
It allows to focus on learning the smaller transformed GM, instead of the original larger one.
On the other hand, our conditioning concept removes some dependencies among variables of GM, 
simply by conditioning some subset of variables.
Hence, it enables us to discover marginalizable structures which was not before conditioning.
At first glance, conditioning looks very powerful as conditioning more variables would discover more desired marginalizable structures.
However, as more variables are conditioned, the algorithmic complexity grows exponentially.
Therefore, we set an upper bound of those conditioned variables.

Marginalization and conditioning
naturally motivate a sequential scheme that repeatedly
recover larger portions of unobserved marginals given previous recovered/ observed ones, i.e.,
recursively
recovering unobserved marginals utilizing any `black-box' local algorithms. 
Developing new local algorithms, other than known factorization methods and exclusive views,
are not of major scope. Nevertheless, we provide two new such algorithms, coined
{disjoint views} and {linear views}, which play a similar role to exclusive views, i.e.,
can also be combined
with known factorization methods.
Given these local algorithms, the proposed sequential learning scheme  can learn a significantly
broader and more complicated class of latent GMs, than known ones, including convolutional restricted Boltzmann machines
and GMs on random regular graphs, as described in Section \ref{sec:example}.
Consequently, our results imply that there exists a one-to-one correspondence between
observed distributions and parameters for the class of latent GMs. 
Furthermore, for arbitrary latent GMs, it can be used for boosting the performance of EM as a pre-processing stage:
first run it to recover as large unobserved marginals as possible, and then run EM using the additional information.
We believe that 
our approach provides a new angle for the important problem of learning latent GMs.



\vspace{0.05in}
\noindent {\bf Related works.}
Parameter estimation of latent GMs has a long history, dating back to \cite{dempster1977maximum}. While it can be broadly applied to most of latent GMs, EM algorithm suffers not only from local optima but from a risk of slow convergence. A natural alternative to \emph{general method} of EM is to constrain the structure of graphical models. In independent component analysis (ICA) and its extensions \cite{Hyvarinen_2000,bach2002kernel}, latent variables are assumed to be independent inducing simple form of latent distribution using products. 
Recently, spectral methods has been successfully applied for various classes of GMs including latent tree \cite{mossel2005learning,song2011kernel}, ICA \cite{comon2010handbook,podosinnikova2015rethinking}, Gaussian mixture models \cite{Hsu2013}, hidden Markov models \cite{siddiqi2010reduced,song2010hilbert,hsu2012spectral,anandkumar2012method,zhang2015spectral}, latent Dirichlet allocation \cite{Anandkumar2012} and others \cite{Halpern2013,Chaganty2013,zou2013contrastive,song2014nonparametric}.
In particular \cite{anandkumar2014tensor} proposed an algorithm of tensor type
under certain graph structures.

Another important line of work using method of moments for latent GMs, concerns on recovering joint or conditional probabilities only among observable variables (see \cite{Balle2012} and its references). \cite{Parikh2011,Parikh2012} proposed spectral algorithms to recover the joint among observable variables when the graph structure is bottlenecked tree. \cite{chaganty2014estimating} relaxed the constraint of tree structure and proposed a technique to combine method of moments in conjunction with likelihood for certain structures. Our generic sequential learning framework allows to use of all these approaches as key components, in order to broaden the applicability of methods. 
We note that we primarily focus on undirected pairwise binary GMs in this paper, but our results can be
naturally extended for other GMs.

%% file: preliminary_arxiv.tex
\vspace{-0.05in}
\section{Preliminaries}\label{sec:pre}
\subsection{Graphical Model and Parameter Learning} 

Given undirected graph $G=(V,E)$, 
we consider the following pairwise binary Graphical Model (GM), where
the joint probability distribution on $x=[x_{i} \in\{0,1\}: i \in V]$ is defined as:
\begin{equation}\label{eq:ising}
\mathbb{P}(x)=
\mathbb{P}_{\beta,\gamma}(x)=\frac{1}{Z}\exp\left(\sum_{(i,j)\in E}\beta_{ij}x_ix_j+\sum_{i\in V}\gamma_ix_i\right),
\end{equation}
for some parameter
$\beta=[\beta_{ij}:(i,j)\in E]\in\mathbb{R}^{E}$ and
$\gamma=[\gamma_i:i\in V]\in\mathbb{R}^{V}$.
The normalization constant $Z$
is called the {\it partition function}.

Given samples $x^{(1)}, x^{(2)}, \cdots,x^{(N)}\in\{0,1\}^V$
drawn
from the distribution \eqref{eq:ising} with
some true (fixed but unknown) parameter $\beta^*,\gamma^*$,
the problem of our interest is recovering it. 
The popular method for the parameter learning task is
the following maximum likelihood estimation (MLE):
\begin{equation}\label{eq:ml}
\mbox{maximize}_{\beta,\gamma}\frac1{N}\sum_{n=1}^N \log \mathbb{P}_{\beta,\gamma}\left(x^{(n)}\right),
\end{equation}
where it is well known \cite{wainwright2008graphical} that the log-likelihood 
$\log \mathbb{P}_{\beta,\gamma}\left(\cdot\right)$ is concave
with respect to $\beta,\gamma$, and the gradient of the log-likelihood is
\begin{align}
&\frac{\partial}{\partial\gamma_i}\frac{1}{N}\sum_{n=1}^N\log \mathbb{P}_{\beta,\gamma}\left(x^{(n)}\right)=
\frac{1}{N}\sum_{n=1}^Nx_i^{(n)}-\mathbb{E}_{\beta,\gamma}[x_i]
\label{eq:gradient1}\\
&\frac{\partial}{\partial\beta_{ij}}\frac{1}{N}\sum_{n=1}^N\log \mathbb{P}_{\beta,\gamma}\left(x^{(n)}\right)=
\frac{1}{N}\sum_{n=1}^N x_i^{(n)}x_j^{(n)}-\mathbb{E}_{\beta,\gamma}[x_ix_j].
\label{eq:gradient2}
\end{align}
Here, the last term, expectation of corresponding sufficient statistics, comes from the partial derivative of the log-partition function. 
Furthermore, it is well known that there exists a one-to-one correspondence between parameter $\beta,\gamma$
and sufficient statistics $\mathbb{E}_{\beta,\gamma}[x_ix_j], \mathbb{E}_{\beta,\gamma}[x_i]$
(see \cite{wainwright2008graphical} for details).

One can further observe that if the number of samples is sufficiently large, i.e., $N\to\infty$,
then \eqref{eq:ml} is equivalent to 
\begin{align*}
       & \mbox{maximize}_{\beta,\gamma}\sum_{x\in\{0,1\}^V}\mathbb{P}_{\beta^*,\gamma^*}(x)\log \mathbb{P}_{\beta,\gamma}(x),
\end{align*}
where 
the true parameter $\beta^*,\gamma^*$ achieves the (unique) optimal solution.
This directly implies that, once empirical nodewise and pairwise marginals in 
\eqref{eq:gradient1}  and \eqref{eq:gradient2} approach the true marginals, 
the gradient method can recover $\beta^*,\gamma^*$ modulo the difficulty of exactly computing the expectations of sufficient statistics.

Now let us consider more challenging task: parameter learning under latent variables.
Given a subset $H$ of $V$ and $O=V\setminus H$,
we assume that for every sample $x=(x_O,x_H)$, 
$x_{O}=\left[x_i\in\{0,1\}: i\in O\right]$
are observed/visible and other variables $x_{H}=\left[x_i\in\{0,1\}: i\in H\right]$ are hidden/latent.
In this case, MLE only involves observed variables: 
\begin{equation}\label{eq:latentml}
\mbox{maximize}_{\beta,\gamma}\frac1N\sum_{n=1}^N \log \mathbb{P}_{\beta,\gamma}\left(x_O^{(n)}\right),
\end{equation}
where $\mathbb{P}_{\beta,\gamma}(x_O)=\sum_{x_H\in \{0,1\}^H} \mathbb{P}_{\beta,\gamma}(x_O, x_H)$.
Similarly as before,
the true parameter $\beta^*,\gamma^*$ achieves the optimal solution of \eqref{eq:latentml}
if the number of samples
is large enough.
However, the log-likelihood under latent variables is no longer concave, which
makes the parameter learning task harder. One can apply an expectation-maximization (EM) scheme, but it is typically stuck in local optima.

\vspace{-0.05in}
\subsection{Tensor Decomposition}\label{sec:tensor}

The fundamental issue on parameter learning of
latent GM is that it is hard to infer 
the pairwise marginals for latent variables, directly from samples.
If one could infer them, it is also possible to recover $\beta^*,\gamma^*$ as we discussed in previous section.
Somewhat surprisingly, however,
under certain conditions of latent GM, pairwise marginals including latent variables can be recovered using low-order visible marginals.
Before introducing such conditions, we first make the following assumption for any GM on a graph $G=(V,E)$ considered throughout
this paper.
\begin{assumption}[Faithful]\label{assum1}
For any two nodes $i,j\in V$, if $i,j$ are connected, then $x_i,x_j$ are dependent.
\end{assumption}
This faithfulness assumption implies that
GM only has conditional independences given by the graph $G$.
We also introduce the following notion \cite{anandkumar2014tensor}. 

\begin{definition}
[Bottleneck] A node $i\in V$ is a bottleneck if there exists $j,k,\ell\in V$, denoted as `views', such that every path between two of $j,k,\ell$ contains $i$.
\end{definition}
\begin{figure}[ht]
\vspace{-0.2in}
\centering
\begin{subfigure}[b]{0.25\textwidth}
\centering
\includegraphics[width=0.8\textwidth]{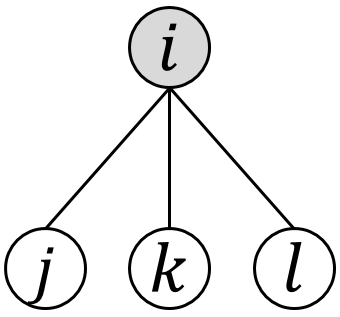}
\caption{bottleneck}
\label{fig:bottleneck}
\end{subfigure}
\qquad\qquad
\begin{subfigure}[b]{0.25\textwidth}
\centering
\includegraphics[width=\textwidth]{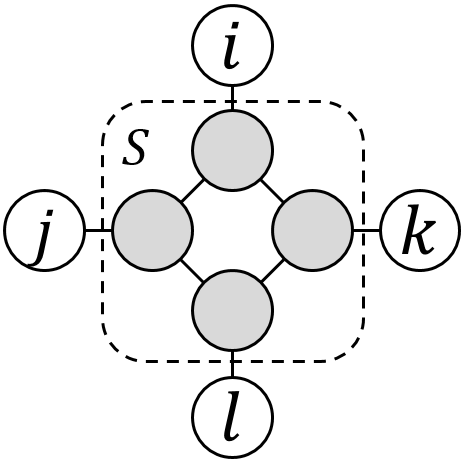}
\caption{exclusive view}
\label{fig:exclusiveview}
\end{subfigure}
\caption{(a) Bottleneck $i$ has three views $j,k,\ell$; (b) Set $S$ satisfies exclusive views property with exclusive views $i,j,k,\ell$}
\vspace{-0.05in}
\end{figure}

Figure \ref{fig:bottleneck} illustrates the bottleneck. By construction, views are conditionally independent given the bottleneck.
Armed with this notion, now we introduce the following theorem to provide sufficient conditions for recovering unobserved/latent marginals \cite{anandkumar2014tensor}. 
\begin{theorem}\label{thm:bottleneck}
Given GM with a parameter $\beta,\gamma$, suppose $i$ is a bottleneck with views $j,k,\ell$.
If $\mathbb{P}_{\beta,\gamma}\left(x_{\{j,k,\ell\}}\right)$ is given,
then there exists an algorithm $\mathtt{TensorDecomp}$ which outputs $\mathbb{P}_{\beta,\gamma}\left(x_{\{i,j,k,\ell\}}\right)$
up to relabeling of $x_i$, i.e. ignoring symmetry of $x_i=0$ and $x_i=1$.
\end{theorem}

The above theorem implies that using visible marginals $\mathbb{P}_{\beta,\gamma}\left(x_{\{j,k,\ell\}}\right)$, one can recover unobserved marginals
$\mathbb{P}_{\beta,\gamma}\left(x_{\{i,j,k,\ell\}}\right)$ involving $x_i$.
For a bottleneck with more than three views, the joint distribution of the bottleneck and views are recoverable using Theorem \ref{thm:bottleneck} by choosing three views at once.

Besides $\mathtt{TensorDecomp}$, there are other conditions of latent GM which marginals including latent variables are recoverable.
Before elaborating on the conditions, we further introduce the following notion for GM on a graph $G=(V,E)$
\cite{chaganty2014estimating}.

\begin{definition}
[Exclusive View] For a set of nodes $S\subset V$,
we say it satisfies the exclusive view property if for each $i\in S$, there exists $j\in V\setminus S$, denoted as `exclusive view', such that every path between $j$ and $S\setminus\{i\}$ contains $i$.
\end{definition}

Figure \ref{fig:exclusiveview} illustrates the exclusive view property.
Now, we are ready to state the conditions for recovering unobserved marginals using the property \cite{chaganty2014estimating}.

\begin{theorem}\label{thm:exclusiveview}
Given GM with a parameter $\beta,\gamma$, suppose a set of nodes $S$ satisfies the exclusive view property with a set of exclusive views $E$.
If $\mathbb{P}_{\beta,\gamma}(x_E)$ and $\mathbb{P}_{\beta,\gamma}\left(x_i,x_j\right)$ are given for all $i\in S$ and an exclusive view $j\in E$ of $i$,
 then there exists an algorithm $\mathtt{ExclusiveView}$ which outputs $\mathbb{P}_{\beta,\gamma}(x_{S\cup E})$.
\end{theorem}

At first glance, Theorem \ref{thm:exclusiveview} does not seems to be useful as it requires a set of marginals including every variable corresponding to $S\cup E$.
However, 
suppose a set of latent nodes $S$ satisfying the property while its set of exclusive views $E$ is visible, i.e., $\mathbb{P}_{\beta,\gamma}(x_E)$ is observed.
If for all $i\in S$, $i$ is a bottleneck with views containing its exclusive view $j\in E$, then one can resort to $\mathtt{TensorDecomp}$ to obtain $\mathbb{P}_{\beta,\gamma}(x_i,x_j)$.

%% file: mainresult1_arxiv.tex
\section{Marginalizing and Conditioning}\label{sec:cond}

In Section \ref{sec:tensor}, we introduced sufficient conditions for recovering unobservable marginals. Specifically, Theorem \ref{thm:bottleneck} and \ref{thm:exclusiveview}
state that for certain structures of latent GMs, it is possible to recover latent marginals simply from low-order visible marginals and in turn the parameters of latent GMs via convex MLE estimators in \eqref{eq:ml}.  

Now, a natural question arises: ``Can we even recover unobserved marginals for latent GMs with more complicated structures beyond naive applications of the bottlenecks or exclusive views?'' To address this, in this section we enlarge the class of such latent GMs by proposing generic 
concepts, marginalization and conditioning. 
\subsection{Key Ideas}
We start by defining two concepts, marginalization and conditioning, formally.
The former is a combinatorial concept defined as follows.
\begin{definition}[Marginalization]\label{def:marginalizable}
 Given graph $G=(V,E)$, we say $S\subset V$ is marginalizable if for all $i\in V\setminus S$, there exists a (minimal) set $S_i\subset S$ with $|S_i|\le 2$ such that $i$ and $S\setminus S_i$ are disconnected in $G\setminus S_i$.\footnote{$G\setminus S_i$
is the subgraph of $G=(V,E)$ induced by $V\setminus S_i$.}
For marginalizable set $S$ in $G=(V,E)$, 
the marginalization of $S$, denoted by $\mathtt{Marg}(S,G)$, 
is the graph on $S$ with edges 
$$\big\{(i,j)\in E\, :\, i,j\in S\big\}\cup\big\{(j,k)\, :\, S_i=\{j,k\} \text{ for } i\in V\setminus S\big\}.$$
\end{definition}

\begin{figure}[ht]
\centering
\begin{subfigure}[b]{0.3\textwidth}
\centering
\includegraphics[width=\textwidth]{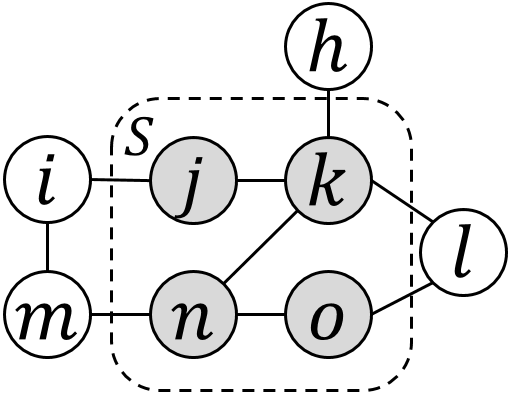}
\caption{$G$ and $S$}
\label{fig:marginalizable1}
\end{subfigure}
\qquad\quad
\begin{subfigure}[b]{0.25\textwidth}
\centering
\includegraphics[width=0.55\textwidth]{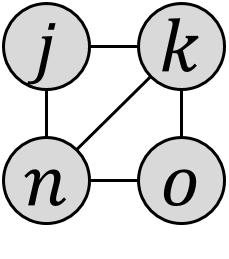}
\caption{$\mathtt{Marg}(S,G)$}
\label{fig:marginalizable2}
\end{subfigure}
\caption{Examples of (a) a graph $G$ and a marginalizable set $S$ in $G$; (b) the marginalization $\mathtt{Marg}(S,G)$ of $S$.}
\label{fig:marginalizable}
\end{figure}

In Figure \ref{fig:marginalizable}, for example, node $i$ is disconnected with $\{k,o\}$ when removing $S_i = \{j,n\}$. Hence, the edge between $j$ and $n$ is additionally included in the marginalization of $S$. 


With the definition of marginalization, the following key proposition reveals that recovering unobserved marginals of a latent GM can be actually reduced to that of much smaller latent GM.

\begin{proposition}\label{lem:marginalizable}
Consider a GM on $G=(V,E)$ with a parameter $\beta,\gamma$.
If $S\subset V$ is marginalizable in $G$, then there exists 
(unique) $\beta^\prime,\gamma^\prime$ such that
GM on $\mathtt{Marg}(S,G)$ with a parameter $\beta^\prime,\gamma^\prime$ inducing
the same distribution on $x_S$, i.e., 
\begin{equation}\label{eq:equalprobGandMarg}
    \mathbb{P}_{\beta,\gamma}(x_S)=\mathbb{P}_{\beta^\prime,\gamma^\prime}(x_S).
\end{equation}
\end{proposition}
The proof of the above proposition is presented in Appendix \ref{sec:pflem:marginalizable}.
Proposition \ref{lem:marginalizable} indeed provides a way of representing the marginal probability on $S$
of GM via the smaller GM on $\mathtt{Marg}(S,G)$. Suppose there exists any algorithm  (e.g., via bottleneck, but we don't restrict ourselves on this method) that can recover a joint distribution $\mathbb{P}_{\beta^\dagger,\gamma^\dagger}(x_S)$, or equivalently
sufficient statistics, of latent GM on $\mathtt{Marg}(S,G)$ only using \emph{observed} marginals in $S$.
Then, it should be 
\begin{equation}\label{eq:equalprobonS}
\mathbb{P}_{\beta^\dagger,\gamma^\dagger}(x_S)=
\mathbb{P}_{\beta^\prime,\gamma^\prime}(x_S),\footnote{Equivalently, $\beta^\dagger=\beta^\prime, \gamma^\dagger=\gamma^\prime$.}
\end{equation} 
where 
$\beta^\prime,\gamma^\prime$ is the unique parameter satisfying \eqref{eq:equalprobGandMarg}. 
Using Proposition \ref{lem:marginalizable} and  marginalization, one can recover unobserved marginals of a large GM by considering smaller GMs corresponding to marginalizations of the large one.
The role of marginalization will be further discussed and clarified in Section \ref{sec:main}.



\begin{figure}[t]
\centering
\begin{subfigure}[b]{0.21\textwidth}
\centering
\includegraphics[width=\textwidth]{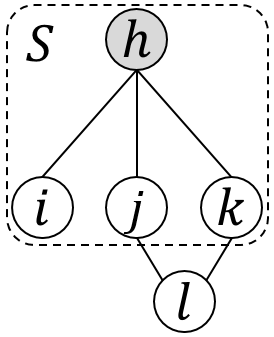}
\caption{$G$ and $S$}
\label{fig:labeling1}
\end{subfigure}
\qquad
\begin{subfigure}[b]{0.21\textwidth}
\centering
\includegraphics[width=\textwidth]{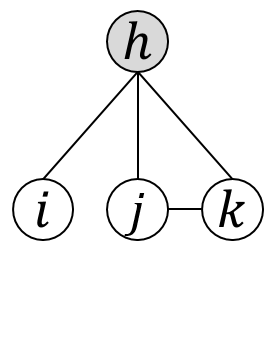}
\caption{$\mathtt{Marg}(S,G)$}
\label{fig:labeling2}
\end{subfigure}
\qquad
\begin{subfigure}[b]{0.21\textwidth}
\centering
\includegraphics[width=\textwidth]{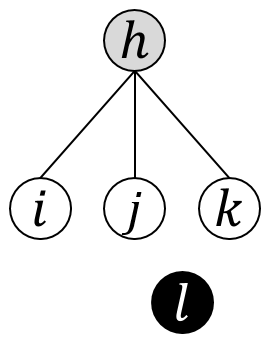}
\caption{$G\setminus\{\ell\}$}
\label{fig:labeling3}
\end{subfigure}
\caption{(a) a graph $G$ and a marginalizable set $S$ (b) the marginalization $\mathtt{Marg}(S,G)$ (c) the marginalization $\mathtt{Marg}(S,G\setminus\{\ell\})=G\setminus\{\ell\}$}
\label{fig:labeling}
\end{figure}

In addition to marginalizing, we introduce the second key ingredient,
called conditioning, with which the class of recoverable latent GMs can be further expanded. 
\begin{proposition}\label{prop:conditioning}
For a graph $G=(V,E)$, for $C\subset V$ and $S\subset V\setminus C$, $\mathtt{Marg}(S,G\setminus C)$ is a subgraph of $\mathtt{Marg}(S,G)$.
\end{proposition}
The proof of the above proposition is straightforward 
since $S_i$ (defined in Definition \ref{def:marginalizable})
for $S$ in $G$ contains that for $S$ in $G\setminus C$, i.e., the edge set of $\mathtt{Marg}(S,G)$ contains that of $\mathtt{Marg}(S,G\setminus C)$.
Figure \ref{fig:labeling} illustrates the example on how conditioning actually broaden the recoverable latent GMs, as suggested in Proposition \ref{prop:conditioning}. Once the node $\ell$ is conditioned out, the marginalization $\mathtt{Marg}(S,G\setminus\{\ell\})$ (Figure \ref{fig:labeling3}) is a form that can be handled by $\mathtt{TensorDecomp}$.  

\subsection{Labeling Issues}\label{sec:label}
In spite of its usefulness, there is a caveat in performing conditioning: consistent labeling of latent nodes.  
For example, consider the latent GM as in Figure \ref{fig:labeling}. Conditioned on $x_\ell$, $h$ is a bottleneck with views $i$, $j$, $k$ (Figure \ref{fig:labeling3}).
If $\mathbb{P}_{\beta,\gamma}\left(x_{\{i,j,k,\ell\}}\right)$ is given,  
one can recover the conditional distribution $\mathbb{P}_{\beta,\gamma}\left(x_{\{h,i,j,k\}}|x_\ell=s\right)$ up to labeling of $x_h$, from
Theorem \ref{thm:bottleneck} and conditioning.
Here, the conditioning worsens the relabeling problem in the sense that we might choose different labels for $x_h$ for each conditioned value $x_\ell=0$ and $x_\ell=1$. As a result, the recovered joint distribution computed as $\sum_{x_\ell\in\{0,1\}}\mathbb{P}_{\beta,\gamma}\left(x_{\{h,i,j,k\}}|x_\ell\right)\mathbb{P}_{\beta,\gamma}(x_\ell)$ with \emph{mixed} labeling of $x_h$, would be different from the true joint.   
To handle this issue, 
we define the following concept for consistent labeling of latent variables.

\begin{definition}
[Label-Consistency] Given GM on $G=(V,E)$ with a parameter $\beta,\gamma$, we say $i\in V$ is label-consistent for $C\subset V\setminus\{i\}$ if 
there exists $j\in V\setminus(C\cup\{i\})$, called `reference', 
such that 
$$\log \frac{\mathbb{P}_{\beta,\gamma}(x_j=1|x_i=1,x_C=s)}{\mathbb{P}_{\beta,\gamma}(x_j=1|x_i=0,x_C=s)},$$
called `preference', is consistently positive or negative
for all $s
\in\{0,1\}^C$.\footnote{Note that the preference cannot be zero due to Assumption \ref{assum1}.}
\end{definition}
In Figure \ref{fig:labeling} for example, $h$ is label-consistent for $\{\ell\}$ with reference $i$ since the corresponding preference is the function only on $\beta_{hi}$, which is fixed as either $\beta_{hi}>0$ or $\beta_{hi}<0$ (note that the reference can be arbitrarily chosen due to the symmetry of structure).
Using the label-consistency of $h$, one can choose a consistent label of $x_h$ by choosing the label consistent to the preference of the reference node $i$.

Even if $i\in V$ is label-consistent under GM with the true known parameter, we need to specify the reference and corresponding preference
to obtain a correct labeling on $x_i$. 
We note however that attractive GMs (i.e., $\beta_{ij}>0$ for all $(i,j)\in E$) always satisfy the label-consistency with any reference node since for any $i,j\in V$ and $C\subset V\setminus\{i,j\}$ where $i,j$ are connected in $G\setminus C$,
$$\mathbb{P}_{\beta,\gamma}(x_j=1|x_i=1,x_C)>\mathbb{P}_{\beta,\gamma}(x_j=1|x_i=0,x_C).$$
Furthermore, there can be some settings in which we can force the label-consistency from the structure of latent GMs even without the information of its true parameter.
For example, consider a latent GM on $G=(V,E)$ and a parameter $\beta,\gamma$.
For a set $C\subset V$, a latent node $i\in V\setminus C$ and its neighbor $j\in V\setminus(C\cup\{i\})$ such that 
$(i,j)\in E$ is the only path from $i$ to $j$ in $G\setminus C$,
by symmetry of labels of latent nodes, one can assume that $\beta_{ij}>0$, i.e., 
$$\mathbb{P}_{\beta,\gamma}(x_j=1|x_i=1,x_C)>\mathbb{P}_{\beta,\gamma}(x_j=1|x_i=0,x_C),$$
to force the label-consistency of $i$ for $C$.
In general, 
one can still choose labels of latent variables to maximize the log-likelihood of observed variables.

As in conditioning, marginalization also has a labeling issue.
Consider a latent GM on $G=(V,E)$.
Suppose that every unobserved pairwise marginal can be recovered by two marginalizations of $S_1,S_2\subset V$.
If there is a common latent node $i\in S_1\cap S_2$, then the labeling for $x_i$ might be inconsistent. 
To address this issue, we make the following assumption on graph $G=(V,E)$, node $i\in V$, and parameter $\beta,\gamma$ of GM.
\begin{assumption}[Degeneracy]\label{assum2}
$\mathbb{P}_{\beta,\gamma}(x_i=1)\ne 0.5$.
\end{assumption}
Under the assumption,
one can choose a label of $x_i$ to satisfy $\mathbb{P}_{\beta,\gamma}(x_i=1)>0.5$ using the symmetry of labels of latent nodes.

%% file: mainresult2_arxiv.tex
\section{Sequential Marginalizing and Conditioning}\label{sec:main}
In the previous section, we introduced two concepts marginalization and conditioning to translate 
the marginal recovery problem of a large GM into that of smaller and tractable GMs.
In this section, we present a sequential strategy, adaptively applying marginalization and conditioning, by which we substantially enlarge the class of tractable GMs with hidden/latent variables.
\subsection{Example}
\begin{figure}[ht]
\centering
\begin{subfigure}[b]{0.2\textwidth}
\centering
\includegraphics[width=\textwidth]{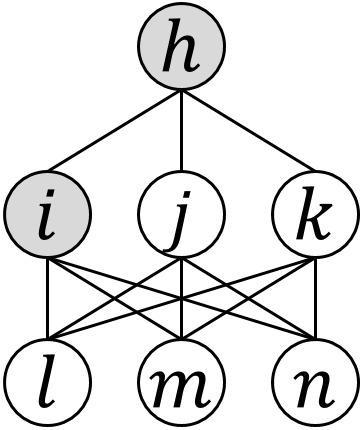}
\caption{}
\label{fig:conditioning1}
\end{subfigure}
\qquad
\begin{subfigure}[b]{0.2\textwidth}
\centering
\includegraphics[width=\textwidth]{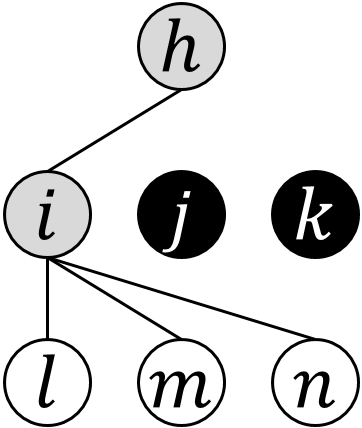}
\caption{}
\label{fig:conditioning2}
\end{subfigure}
\qquad
\begin{subfigure}[b]{0.2\textwidth}
\centering
\includegraphics[width=\textwidth]{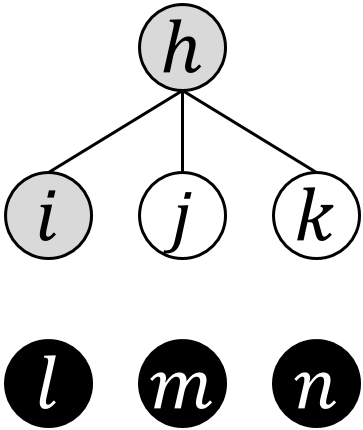}
\caption{}
\label{fig:conditioning3}
\end{subfigure}
\caption{(a) A latent GM with latent nodes $h,i$ and visible nodes $j,k,\ell,m,n$ (b) A latent GM after conditioning $x_{\{j,k\}}$ (c) A latent GM after conditioning $x_{\{\ell,m,n\}}$}
\label{fig:conditioning}
\end{figure}

We begin with a simple example describing our sequential learning framework. Consider a latent GM as illustrated in Figure \ref{fig:conditioning1} and a parameter $\beta,\gamma$.
Given visible marginal $\mathbb{P}_{\beta,\gamma}\left(x_{\{j,k,\ell,m,n\}}\right)$, our goal is to recover all unobserved pairwise marginals including $x_h$ or $x_i$ in order to learn $\beta,\gamma$ via convex MLE \eqref{eq:ml}.
As both nodes $h$ and $i$ are not a bottleneck, one can consider the conditioning strategy described
in the previous section, i.e.,
the conditional distribution $\mathbb{P}_{\beta,\gamma}\left(x_{\{h,i,\ell,m,n\}}|x_{\{j,k\}}\right)$ in Figure \ref{fig:conditioning2}.
Now, node $i$ is a bottleneck with views $\ell,m,n$. 
Hence, one can recover $\mathbb{P}_{\beta,\gamma}\left(x_{\{i,\ell,m,n\}}|x_{\{j,k\}}\right)$ using $\mathtt{TensorDecomp}$ where the label of $x_i$ is set to satisfy 
\begin{align*}
\mathbb{P}_{\beta,\gamma}&\left(x_\ell=1|x_i=1,x_{\{j,k\}}\right) > \mathbb{P}_{\beta,\gamma}\left(x_\ell=1|x_i=0,x_{\{j,k\}}\right),
\end{align*}
i.e., node $i$ is label consistent.
Further, $\mathbb{P}_{\beta,\gamma}\left(x_{\{i,j,k,\ell,m,n\}}\right)$ can be recovered
using the known visible marginals $\mathbb{P}_{\beta,\gamma}\left(x_{\{j,k\}}\right)$ and the following identity
$$\mathbb{P}_{\beta,\gamma}\left(x_{\{i,j,k,\ell,m,n\}}\right)=\mathbb{P}_{\beta,\gamma}\left(x_{\{i,\ell,m,n\}}|x_{\{j,k\}}\right)\mathbb{P}_{\beta,\gamma}\left(x_{\{j,k\}}\right).$$
Since we recovered pairwise marginals between $x_i$ and $x_\ell$, $x_m$, $x_n$, the remaining goal is to recover pairwise marginals including $x_h$.
Now consider a latent GM where $x_{\{\ell,m,n\}}$ is conditioned and it
is illustrated in Figure \ref{fig:conditioning3}.
At this time, the node $h$ is a bottleneck with views $i,j,k$, which can be handled by an additional application of $\mathtt{TensorDecomp}$ (the details are same as the previous case on node $i$). 

This example shows that the sequential application of conditioning extends a class of latent GM that unobserved pairwise marginals are recoverable.
Here, we use an algorithm $\mathtt{TensorDecomp}$ as a black-box, hence one can consider other algorithms as long as they have similar guarantees.
One caveat is that conditioning an arbitrary number of variables is very expensive 
as the learning algorithmic (and sampling) complexity grows exponentially with respect to the number of conditioned variables.
Therefore, it would be reasonable to bound the number of conditioned variables.

\subsection{Algorithm Design}
Now, we are ready to state the main learning framework sequentially applying marginalization and conditioning, summarized in Algorithm \ref{alg:sequential}. 
Suppose that there exists an algorithm, called $\mathtt{NonConvexSolver}$, e.g.,
$\mathtt{TensorDecomp}$, for a class of pairs $\mathcal{N}\subset\{(G,\mathcal{S}_G):G=(V,E),\mathcal{S}_G\subset 2^V\}$ such that  
all $(G,\mathcal{S}_G)\in\mathcal{N}$ satisfy the following:
\begin{itemize}
    \item[$\circ$]
    Given GM with a parameter $\beta,\gamma$ on $G=(V,E)$
    and marginals $\{\mathbb{P}_{\beta,\gamma}\left(x_S\right):S\in\mathcal{S}_G\}$, $\mathtt{NonConvexSolver}$ outputs the entire distribution $\mathbb{P}_{\beta,\gamma}(x)$,
up to labeling of variables on $V\setminus\left(\bigcup_{S\in\mathcal{S}_G}S\right)$.
\end{itemize}
For example, consider a graph $G$ illustrated in Figure \ref{fig:bottleneck}
with $\mathcal{S}_G=\big\{\{j,k,\ell\}\big\}$.
Then, $\mathtt{TensorDecomp}$ outputs the entire distribution $\mathbb{P}_{\beta,\gamma}\left(x_{\{i,j,k,\ell\}}\right)$.

In addition, suppose that there exists an algorithm, called $\mathtt{Merge}$, e.g., $\mathtt{ExclusiveView}$, for a class of pairs $\mathcal{M}\subset\{(G,\mathcal{T}_G):G=(V,E),\mathcal{T}_G\subset 2^V\}$ such that all
$(G,\mathcal{T}_G)$ satisfy the following:
\begin{itemize}
    \item[$\circ$]
    Given GM with a parameter $\beta,\gamma$ on $G=(V,E)$
    and marginals $\{\mathbb{P}_{\beta,\gamma}\left(x_S\right):S\in\mathcal{T}_G\}$, $\mathtt{Merge}$ outputs the distribution $\mathbb{P}_{\beta,\gamma}\left(x_{T}\right)$ where $T=\bigcup_{S\in\mathcal{T}_G}S$.
\end{itemize}
Namely, $\mathtt{Merge}$ simply merges the small marginal distributions for $S\in\mathcal{T}_G$ into 
the entire distribution on $\bigcup_{S\in\mathcal{T}_G}S$.
For example, consider a graph $G$ illustrated in Figure \ref{fig:exclusiveview} with $$\mathcal{T}_G=\big\{\{i,j,k,\ell\},\{i,i^\prime\},\{j,j^\prime\},\{k,k^\prime\},\{\ell,\ell^\prime\}\big\}$$
where $i^\prime,j^\prime,k^\prime,\ell^\prime\in S$ have exclusive views $i,j,k,\ell$, respectively.
Then, $\mathtt{ExclusiveView}$ outputs the distribution $\mathbb{P}_{\beta,\gamma}\left(x_{S\cup\{i,j,k,\ell\}}\right)$.

For a GM on $G=(V,E)$ with a parameter $\beta,\gamma$, suppose we know
a family of label-consistency quadruples
\begin{align*}
\mathcal{L}=\big\{&(i,j,p,C)\,:\,\text{$i$ is label-consistent for $C$}\\
&\qquad\qquad  \text{with reference $j$ and preference $p$}\big\}
\end{align*}
and marginals $\{\mathbb{P}_{\beta,\gamma}(x_S):S\in\sigma_0\}$ for some $\sigma_0\subset 2^V$.
As we mentioned in the previous section, we also bound the number of conditioning variables by some $K\ge 0$.
Under the setting, our goal is to recover 
more marginals beyond initially known
ones $\{\mathbb{P}_{\beta,\gamma}(x_S):S\in\sigma_0\}$.

The following conditions 
for $C\subset V$ with $|C|\le K$ and $R\subset V\setminus C$
are sufficient 
so that additional marginals
$\mathbb{P}_{\beta,\gamma}(x_{R\cup C})$ can be recovered by conditioning variables on $C$, marginalizing $R$ and applying $\mathtt{NonConvexSolver}$:
\begin{itemize}
\item[$\mathcal C1.$] $(H,\mathcal{S}_H)\in\mathcal{N}$ for some $\mathcal{S}_H\subset 2^V$
\item[$\mathcal C2.$] For all $S\in\mathcal{S}_H$, there exists $S^\prime\in\sigma_0$ such that $S\cup C\subset S^\prime$   
\item[$\mathcal C3.$] For all $i\in R\setminus\left(\bigcup_{S\in\mathcal{S}_H}S\right)$, there exist $j\in\bigcup_{S\in\mathcal{S}_H}S$ and $p$ such that $(i,j,p,C)\in\mathcal{L}$,
\end{itemize}
where $H=\mathtt{Marg}(R,G\setminus C)$.
In the above,
$\mathcal{C}1$ implies that if $\{\mathbb{P}_{\beta,\gamma}(x_S|x_{C}):S\in\mathcal{S}_H\}$ are given,
then
$\mathtt{NonConvexSolver}$ outputs $\mathbb{P}_{\beta,\gamma}(x_R|x_{C})$ up to labeling of $R\setminus\left(\bigcup_{S\in\mathcal{S}_H}S\right)$.
In addition,
$\mathcal{C}2$ says that the required marginals $\{\mathbb{P}_{\beta,\gamma}(x_S|x_{C}):
S\in\mathcal{S}_H\}$ and $\mathbb{P}(x_C)$ are known.
Finally, $\mathcal{C}3$ is necessary that all nodes which we need to infer their labels are label-consistent.

Similarly, the following conditions for $C\subset V$ with $|C|\le K$ and $(G\setminus C,\mathcal{T}_{G\setminus C})\in\mathcal{M}$ are sufficient so that $\mathbb{P}_{\beta,\gamma}(x_{T\cup C})$ can be recovered by conditioning variables on $C$ and applying $\mathtt{Merge}$ where $T=\bigcup_{S\in\mathcal{T}_G}S$:
\begin{itemize}
\item[$\mathcal C4.$] For all $S\in\mathcal{T}_{G\setminus C}$, there exists $S^\prime\in\sigma_0$ such that $S\cup C\subset S^\prime$,
\end{itemize}
In the above, 
$\mathcal{C}4$ says that the required marginals for merging are given.

The above procedures imply that
given initial marginals $\{\mathbb{P}_{\beta,\gamma}(x_S):S\in\sigma_0\}$, one can recover \emph{additional} marginals 
 $\{\mathbb{P}_{\beta,\gamma}(x_S):S\in\mathcal{A}_0\cup \mathcal{B}_0\}$, where  
\begin{align}\label{eq:AtBt}
    \mathcal{A}_0=\{&R\cup C:C\subset V,|C|\le K,R\subset V\setminus C\nonumber~\text{satisfy $\mathcal{C}1$-$\mathcal{C}3$}\},\nonumber\\
    \mathcal{B}_0=\{&T\cup C:C\subset V,|C|\le K,(G\setminus C,\mathcal{T}_{G\setminus C})\in\mathcal{M}\nonumber\\
    &\qquad\qquad\text{satisfy $\mathcal{C}4$ where $T=\cup_{S\in\mathcal{T}_G}S$}\},
\end{align}
from $\mathtt{NonConvexSolver}$ and $\mathtt{Merge}$, respectively.
One can repeat the above procedure for recovering more marginals as
$$\sigma_{t+1}=\sigma_t\cup {\mathcal A}_{t}\cup {\mathcal B}_t.$$
Recall that we are primarily interested in recovering 
all pairwise marginals, i.e., $$\{\mathbb{P}_{\beta,\gamma}(x_i,x_j): (i,j)\in E\}.$$
The following theorem implies that one can check the success of Algorithm \ref{alg:sequential} in
$O\left(|V|^{K+L}\right)$ time, where $K,L$ are typically chosen as small constants.
\begin{theorem}\label{thm:main}
Suppose we have a label-consistency family $\mathcal L$ of GM on $G=(V,E)$ and marginals 
$\{\mathbb{P}_{\beta,\gamma}(x_S):S\in\sigma_0\}$ for some $\sigma_0\subset 2^V$.
If Algorithm \ref{alg:sequential} eventually recover all pairwise marginals, 
then they do 
in $O\left(|V|^{K+L}\right)$ iterations,
where $K$ and $L$ denote the maximum numbers of conditioning variables
and nodes of graphs in $\mathcal{N},\mathcal{M}$, respectively.
\end{theorem}
The proof of the above theorem is presented in Appendix \ref{sec:pfthm:main}.
We note that one can design their own sequence of recovering marginals rather than recovering all marginals in $\mathcal{A}_t,\mathcal{B}_t$ for computational efficiency.
In Section \ref{sec:example}, we provide such examples, 
of which strategy has the linear-time complexity at each iteration. 
We also remark that even when Algorithm \ref{alg:sequential} recovers some, not all, pairwise unobserved marginals
for given latent GMs,
it is still useful since one can run the EM algorithm using the additional information provided by Algorithm \ref{alg:sequential}. 
We leave this suggestion for further exploration in the future.

\begin{algorithm}[t]
\caption{Sequential Local Learning} \label{alg:sequential}
\begin{algorithmic}[1]
\STATE {\bf Input } $G=(V,E)$, Initially observable marginals $\{\mathbb{P}_{\beta,\gamma}(x_S):S\in\sigma_0\}$, $\mathtt{NonConvexSolver}$, $\mathtt{Merge}$ 
\WHILE{until convergence} 
\STATE $\sigma_{t+1}=\sigma_t\cup {\mathcal A}_{t}\cup {\mathcal B}_t$ from \eqref{eq:AtBt}
\ENDWHILE
\STATE {\bf Return } All recovered pairwise marginals
\end{algorithmic}
\end{algorithm}

\subsection{Recoverable Local Structures}\label{sec:exalg}
For running the sequential learning framework in the previous section,
one requires `black-box' knowledge of 
a label-consistency family $\mathcal L$ and a class of locally recoverable structures of latent GMs, i.e.,
$\mathcal{N}$ and $\mathcal{M}$.
The complete study on them is out of our scope, but we provide the following guidelines on their choices.

As mentioned in Section \ref{sec:label},
$\mathcal L$ can be found easily for some class of GMs including attractive ones.
One can also infer it heuristically for general GMs in practice.
As we mentioned in the previous section,
one can choose $(G,\mathcal{S}_G)\in\mathcal{N}$ that corresponds to
$\mathtt{TensorDecomp}$.
Beyond $\mathtt{TensorDecomp}$, in practice, one might hope to choose an
additional option for small sized latent GMs since even a generic non-convex solver
might compute an almost optimum of MLE
due to their small dimensionality.

For the choice of $(G,\mathcal{T}_G)\in\mathcal M$,
we mentioned those corresponding to 
$\mathtt{ExclusiveView}$ in the previous section.
In addition, we provide the following two more examples, called
$\mathtt{DisjointView}$ and $\mathtt{LinearView}$, as described in Algorithm \ref{alg:disjointview}
and \ref{alg:linearview}, respectively.
In Algorithm \ref{alg:linearview}, 
$[\mathbb{P}_{\beta,\gamma}(x_j,x_i)]^{-1}$  is defined as 
\begin{align*}
\begin{bmatrix}
\mathbb{P}_{\beta,\gamma}(x_j=0,x_i=0)&\mathbb{P}_{\beta,\gamma}(x_j=0,x_i=1)\\
\mathbb{P}_{\beta,\gamma}(x_j=1,x_i=0)&\mathbb{P}_{\beta,\gamma}(x_j=1,x_i=1)
\end{bmatrix}^{-1}.
\end{align*}

\begin{algorithm}[H]
\caption{$\mathtt{DisjointView}$} \label{alg:disjointview}
\begin{algorithmic}[1]
\STATE {\bf Input } $\mathcal{T}_G=\{S\cup C,T\cup C\}$, $\{\mathbb{P}_{\beta,\gamma}(x_S):S\in\mathcal{T}_G\}$ 
\STATE \qquad\quad$S,T$ are disconnected in $G\setminus C$
\STATE $\mathbb{P}_{\beta,\gamma}(x_{S\cup T\cup C})$
\STATE \qquad $\leftarrow\mathbb{P}_{\beta,\gamma}(x_S|x_C)\mathbb{P}_{\beta,\gamma}(x_T|x_C)\mathbb{P}_{\beta,\gamma}(x_C)$
\STATE {\bf Return } $\mathbb{P}_{\beta,\gamma}(x_{S\cup T\cup C})$
\end{algorithmic}
\end{algorithm}

\begin{algorithm}[H]
\caption{$\mathtt{LinearView}$} \label{alg:linearview}
\begin{algorithmic}[1]
\STATE {\bf Input } $\mathcal{T}_G=\{\{i,j\},S\cup\{j\}\}$, $\{\mathbb{P}_{\beta,\gamma}(x_S):S\in\mathcal{T}_G\}$ 
\STATE\qquad\quad $S,j$ are disconnected in $G\setminus\{i\}$
\FOR{$s\in\{0,1\}^S$}
\STATE $\mathbb{P}_{\beta,\gamma}(x_{S}=s|x_i)$ 
\STATE \qquad
$\leftarrow[\mathbb{P}_{\beta,\gamma}(x_j,x_i)]^{-1}$
$\mathbb{P}_{\beta,\gamma}(x_j,x_{S}=s)$
\ENDFOR
\STATE {\bf Return } $\mathbb{P}_{\beta,\gamma}\left(x_{S\cup\{i,j\}}\right)$
\end{algorithmic}
\end{algorithm}

\begin{figure}[H]
\centering
\begin{subfigure}[b]{0.36\textwidth}
\centering
\includegraphics[width=\textwidth]{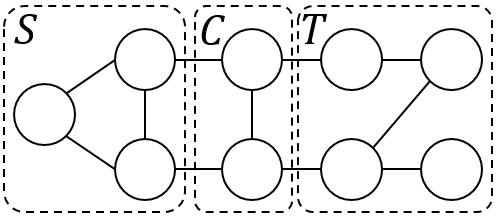}
\caption{$\mathtt{DisjointView}$}
\label{fig:disjointview}
\end{subfigure}
\qquad
\begin{subfigure}[b]{0.29\textwidth}
\centering
\includegraphics[width=\textwidth]{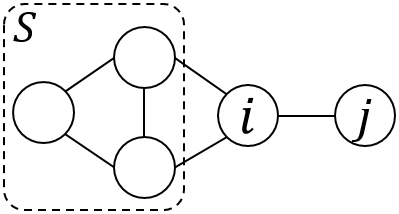}
\caption{$\mathtt{LinearView}$}
\label{fig:linearview}
\end{subfigure}
\caption{Illustrations for (a) $\mathtt{DisjointView}$ (b) $\mathtt{LinearView}$}
\label{fig:merge}
\end{figure}
\noindent Figure 
\ref{fig:merge}
illustrates $\mathtt{DisjointView}$ and $\mathtt{LinearView}$.

%% file: example_arxiv.tex
\section{Examples}\label{sec:example}
In this section, we provide concrete examples of loopy
latent GM where the proposed sequential learning
framework 
is applicable.
In what follows,
we assume that it uses classes $\mathcal N, \mathcal M$
corresponding to $\mathtt{TensorDecomp}$, $\mathtt{ExculsiveView}$, $\mathtt{DisjointView}$ and $\mathtt{LinearView}$.

\begin{figure}[t]
\centering
\begin{subfigure}[b]{0.145\textwidth}
\centering
\includegraphics[width=\textwidth]{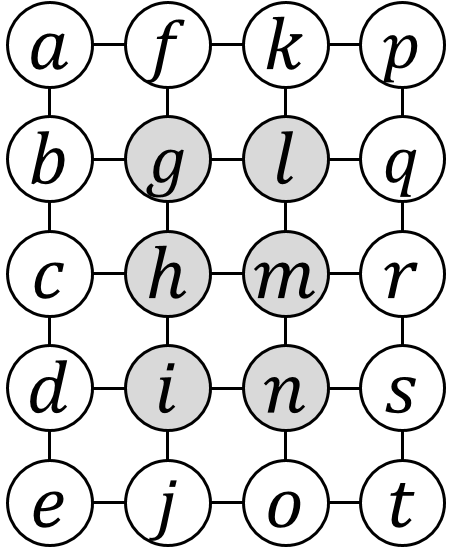}
\caption{}
\label{fig:grid1}
\end{subfigure}
~
\begin{subfigure}[b]{0.145\textwidth}
\centering
\includegraphics[width=\textwidth]{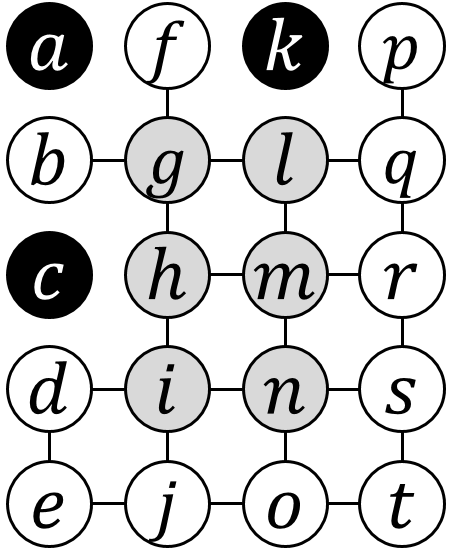}
\caption{}
\label{fig:grid2}
\end{subfigure}
~
\begin{subfigure}[b]{0.145\textwidth}
\centering
\includegraphics[width=\textwidth]{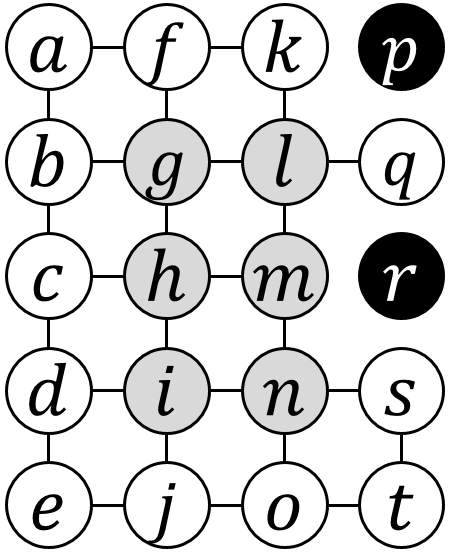}
\caption{}
\label{fig:grid3}
\end{subfigure}
~
\begin{subfigure}[b]{0.145\textwidth}
\centering
\includegraphics[width=\textwidth]{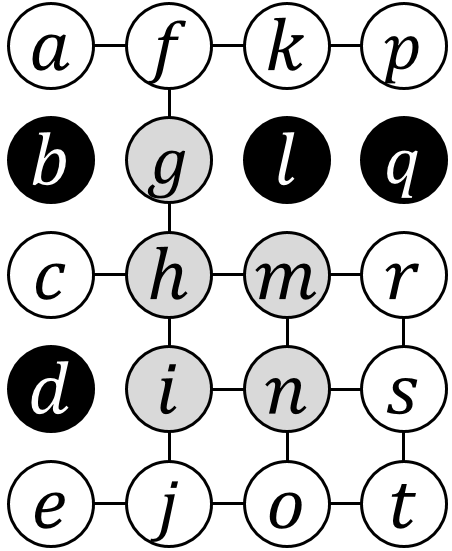}
\caption{}
\label{fig:grid4}
\end{subfigure}
~
\begin{subfigure}[b]{0.145\textwidth}
\centering
\includegraphics[width=\textwidth]{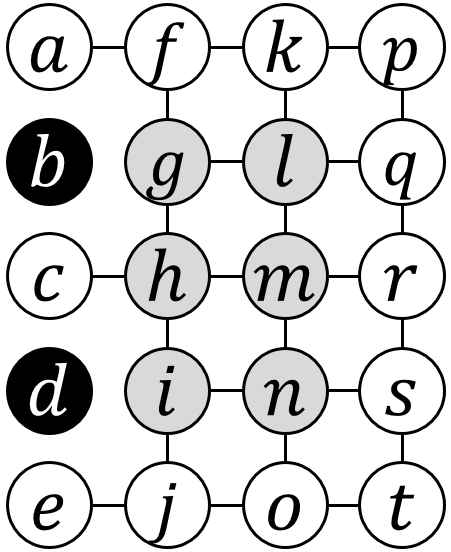}
\caption{}
\label{fig:grid5}
\end{subfigure}
~
\begin{subfigure}[b]{0.145\textwidth}
\centering
\includegraphics[width=\textwidth]{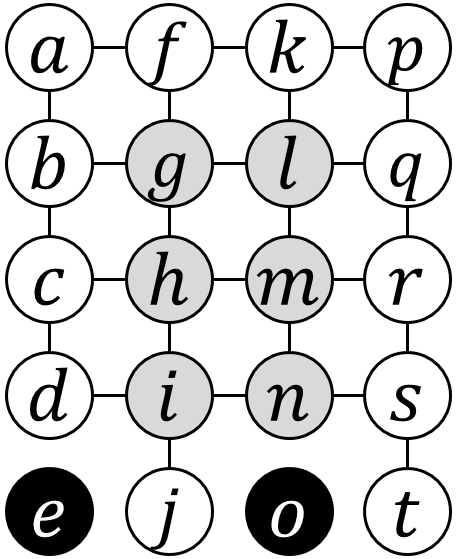}
\caption{}
\label{fig:grid6}
\end{subfigure}

\caption{Sequential learning for recovering $\mathbb{P}_{\beta,\gamma}(x_h,x_i)$ (a) GM on a grid graph (b) Recover $\mathbb{P}_{\beta,\gamma}\left(x_{\{a,b,c,f,g,k,p\}}\right)$ using that $g$ is a bottleneck with views $b,f,p$ conditioned on $x_{\{a,c,k\}}$. Similarly, recover $\mathbb{P}_{\beta,\gamma}\left(x_{\{a,f,k,\ell,p,q,r\}}\right)$ and $\mathbb{P}_{\beta,\gamma}\left(x_{\{c,d,e,i,j,o,t\}}\right)$ (c) Recover $\mathbb{P}_{\beta,\gamma}\left(x_{\{b,c,d,\ell,p,q,r\}}\right)$ using that $S\leftarrow\{b,c,d\}$, $i\leftarrow\ell$ and $j\leftarrow q$ form an input of $\mathtt{LinearView}$ conditioned on $x_{\{p,q\}}$ (d) Recover $\mathbb{P}_{\beta,\gamma}\left(x_{\{b,c,d,h,\ell,p,q,r\}}\right)$ using that $h$ is a bottleneck with views $c,p,r$ conditioned on $x_{\{b,d,\ell,q\}}$ (e) Recover $\mathbb{P}_{\beta,\gamma}\left(x_{\{b,c,d,h,e,j,o\}}\right)$ using that $S\leftarrow\{e,j,o\}$, $i\leftarrow h$ and $j\leftarrow c$ form an input of $\mathtt{LinearView}$ conditioned on $x_{\{b,d\}}$ (f) Recover $\mathbb{P}_{\beta,\gamma}\left(x_{\{e,h,i,j,o\}}\right)$ using that $S\leftarrow\{h\}$, $i\leftarrow i$ and $j\leftarrow j$ form an input of $\mathtt{LinearView}$ conditioned on $x_{\{e,o\}}$} 
\label{fig:grid}
\end{figure}

\vspace{0.05in}
\noindent {\bf Grid graph.}
We first consider a latent GM on a grid graph illustrated in Figure \ref{fig:grid1} where boundary nodes are visible and internal nodes are latent.
The following lemma states that 
all pairwise marginals
can be successfully recovered given observed ones, utilizing the proposed sequential learning algorithm.

\begin{lemma}\label{thm:grid}
Consider any latent GM with a parameter $\beta,\gamma$ illustrated in Figure \ref{fig:grid1}, $K=3$,
and $\sigma_0=\{S\subset O:|S|\le 6\}$. 
Then, $\sigma_{5}$ updated under Algorithm \ref{alg:sequential} contains all pairwise marginals.
\end{lemma}

In the above, recall that
$O$ is the set of visible nodes.
The proof strategy is illustrated in Figure \ref{fig:grid}
and the formal proof is presented in Appendix \ref{sec:pflem:grid}. 
We remark that to prove Lemma \ref{thm:grid}, $\mathtt{ExclusiveView}$ and $\mathtt{DisjointView}$ are not necessary to use.

\begin{figure}[t]
\centering
\begin{subfigure}[b]{0.21\textwidth}
\centering
\includegraphics[width=\textwidth]{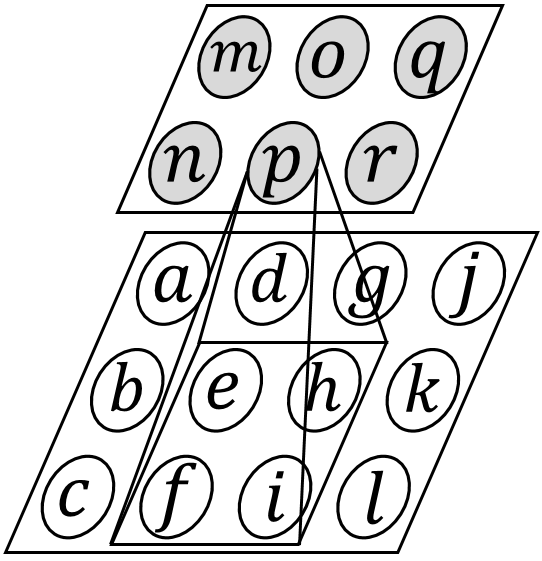}
\caption{}
\label{fig:crbm1}
\end{subfigure}
\qquad
\begin{subfigure}[b]{0.21\textwidth}
\centering
\includegraphics[width=\textwidth]{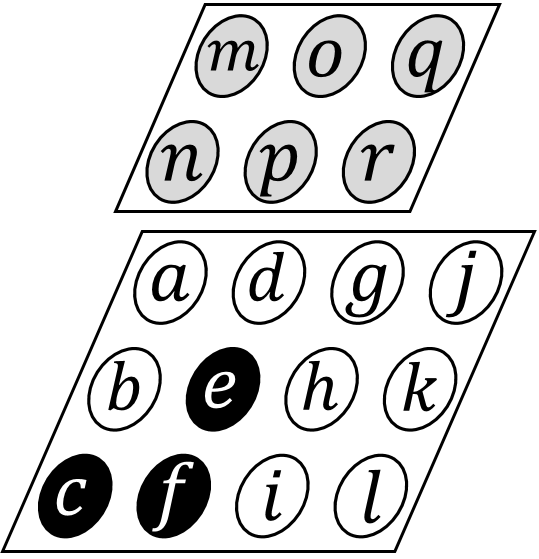}
\caption{}
\label{fig:crbm2}
\end{subfigure}
\qquad
\begin{subfigure}[b]{0.21\textwidth}
\centering
\includegraphics[width=\textwidth]{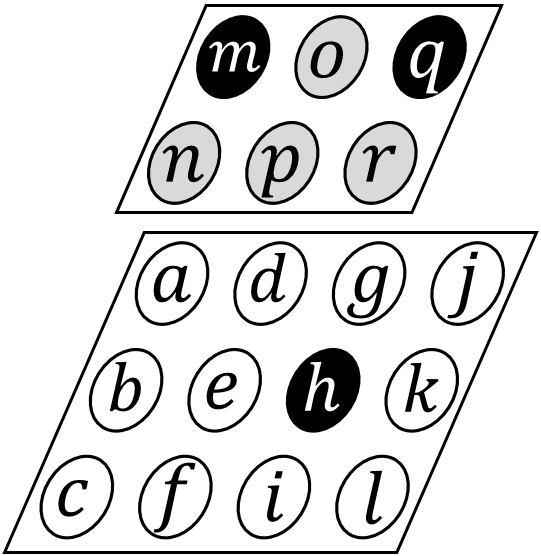}
\caption{}
\label{fig:crbm3}
\end{subfigure}

\caption{Sequential learning for recovering $\mathbb{P}_{\beta,\gamma}\left(x_{\{d,e,g,h,o\}}\right)$ (a)  CRBM where edges exist between $m$ and $\{a,b,d,e\}$, $n$ and $\{b,c,e,f\}$, $o$ and $\{d,e,g,h\}$, $p$ and $\{e,f,h,i\}$, $q$ and $\{g,h,j,k\}$, $r$ and $\{h,i,k,\ell\}$ (b) Recover $\mathbb{P}_{\beta,\gamma}\left(x_{\{a,b,c,d,e,f,m\}}\right)$ using that $m$ is a bottleneck with views $a,b,d$ conditioned on $x_{\{c,e,f\}}$. Similarly, recover $\mathbb{P}_{\beta,\gamma}\left(x_{\{g,h,i,j,k,\ell,q\}}\right)$ (c) Recover $\mathbb{P}_{\beta,\gamma}\left(x_{\{a,b,d,e,g,h,j,k,m,q\}}\right)$ using $\mathtt{DisjointView}$.
Then, Recover $\mathbb{P}_{\beta,\gamma}\left(x_{\{d,e,g,h,m,o,q\}}\right)$ using that $o$ is a bottleneck with views $d,e,g$ conditioned on $x_{\{h,m,q\}}$} 
\label{fig:crbm}
\end{figure}

\vspace{0.05in}
\noindent {\bf Convolutional graph.}
Second,
we consider a latent GM illustrated in Figure \ref{fig:crbm1},
which corresponds to a convolutional restricted Boltzmann machine (CRBM) \cite{lee2009convolutional},
and also prove the following lemma.

\begin{lemma}\label{lem:convRBM}
Consider any latent GM with a parameter $\beta,\gamma$ illustrated in Figure \ref{fig:crbm1}, $K=3$,
and $\sigma_0=\{S\subset O:|S|\le 8\}$. 
Then, $\sigma_{4}$ updated under Algorithm \ref{alg:sequential} contains all pairwise marginals.
\end{lemma}

The proof strategy is illustrated in Figure \ref{fig:crbm}
and the formal proof is presented again in Appendix \ref{sec:pflem:convRBM}. 
We remark that to prove Lemma \ref{lem:convRBM}, $\mathtt{ExclusiveView}$ and $\mathtt{LinearView}$ are not necessary to use.
Furthermore, it is straightforward to generalize the proof of Lemma \ref{lem:convRBM} 
for arbitrary CRBM.
\begin{lemma}\label{thm:convRBM}
Consider any CRBM with $N\times M$ visible nodes and a filter size $n\times m$, $2\le n\le m$, $K=2mn-4$
and $\sigma_0=\{S\subset O:|S|\le 4mn-2m\}$. 
Then, $\sigma_{MNmn/2}$ updated under Algorithm \ref{alg:sequential} contains all pairwise marginals.\footnote{The theorem holds for arbitrary stride of CRBM.}
\end{lemma}

\vspace{0.05in}
\paragraph{\bf Random regular graph.}
Finally, we state the following theorem for latent random regular GMs.
\begin{lemma}\label{thm:regular}
Consider any latent GM with a parameter $\beta,\gamma$
on a random $d$-regular graph $(V,E)$ for some constant $d\geq 5$, $K=2d-2$ and
$\sigma_0=\{S\subset O:|S|\le 2(d-1)^2|H|)\}$.
There exists a constant $c=c(d)$ such that if the number of latent variables is at most $c=c|V|$,
$\sigma_{2d|H|}$ updated under Algorithm \ref{alg:sequential} contains all pairwise marginals a.a.s.
\end{lemma}
The proof of the above lemma is presented in Appendix \ref{sec:pflem:regular},
where it is impossible without using our sequential learning strategy.
One can obtain an explicit formula of $c(d)$ from our proof, but it is quite a loose bound since we do not make much efforts to optimize it.

%% file: conclusion_arxiv.tex
\section{Conclusion}
In this paper, we present a new learning strategy for latent graphical models.
Unlike known algebraic, e.g., $\mathtt{TensorDecomp}$ and optimization, e.g., 
$\mathtt{ExculsiveView}$, approaches for this non-convex problem,
ours is of combinatorial flavor and more generic using them as subroutines.
We believe that our approach provides a new angle for the important learning task. 

%% file: appendix_arxiv.tex
\appendix
\clearpage
\onecolumn



\section{Proof of Proposition \ref{lem:marginalizable}}\label{sec:pflem:marginalizable}
We use the mathematical induction on $|\{S_i:i\in V\setminus S\}|$ where $S_i$ is defined in Definition \ref{def:marginalizable}.
Before starting the proof we define the equivalence class $[\ell]=\{i\in V\setminus S:S_i=S_\ell\}$.
Now, we start the proof by considering
\begin{align*}
&\sum_{i\in V\setminus S}\mathbb{P}_{\beta,\gamma}(x)=\sum_{i\in V\setminus S:i\notin[\ell]}\sum_{i\in V\setminus S:i\in[\ell]}\frac1Z\exp\left(\sum_{(i,j)\in E}\beta_{ij}x_ix_j+\sum_{i\in V}\gamma_ix_i\right)\\
&=\sum_{i\in V\setminus S:i\notin[\ell]}\frac1Z\exp\left(\sum_{(i,j)\in E:i,j\notin[\ell]}\beta_{ij}x_ix_j+\sum_{i\in V\setminus[\ell]}\gamma_ix_i\right)\\
&\qquad\times\sum_{i\in V\setminus S:i\in[\ell]}\exp\left(\sum_{(i,j)\in E:i\in[\ell]}\beta_{ij}x_ix_j+\sum_{i\in [\ell]}\gamma_ix_i\right)\\
&=\sum_{i\in V\setminus S:i\notin[\ell]}\frac1Z\exp\left(\sum_{(i,j)\in E:i,j\notin[\ell]}\beta_{ij}x_ix_j+\sum_{i\in V\setminus[\ell]}\gamma_ix_i\right)f_{[\ell]}(x_{S_\ell})
\end{align*}
where $f_{[\ell]}(x_{S_\ell})$ is some positive function.
Since $|S_\ell|\le 2$, one can modify a parameter $\beta^\dagger,\gamma^\dagger$ only between elements of $S_\ell$ to achieve the following identity
$$\sum_{i\in V\setminus S}\mathbb{P}_{\beta,\gamma}(x)=\sum_{i\in V\setminus\{S\cup[\ell]\}}\frac1{Z^\dagger}\exp\left(\sum_{(i,j)\in E^\dagger}\beta_{ij}^\dagger x_ix_j+\sum_{i\in V\setminus[\ell]}\gamma_i^\dagger x_i\right)$$
where $E^\prime=E\cup\{(j,k):S_\ell=\{j,k\}\}$.
Using the induction hypothesis, the above identity completes the proof of Proposition \ref{lem:marginalizable}.

\section{Proof of Theorem \ref{thm:main}}\label{sec:pfthm:main}
Since the algorithm only uses the marginals of at most $K+L$ dimensions, instead of $\sigma_t$, consider the following sequence
$$\sigma^\prime_t=\{S\subset S^\prime:S^\prime\in\sigma_t,|S|\le K+L\}.$$
One can observe that if $\sigma^\prime_t=\sigma^\prime_{t-1}$, then one can observe that the sequential local framework cannot recover more marginals after $t$-th iteration, while $\sigma_t$ increases its cardinality at least $1$ otherwise.
However, the maximum cardinality of $\sigma_t$ is $O(|V|^{K+L})$ and this implies that the algorithm always terminates in $O(|V|^{K+L})$.
This completes the proof of Theorem \ref{thm:main}.

\section{Proof of Lemma \ref{thm:grid}}\label{sec:pflem:grid}

We first consider the distribution conditioned on $x_{\{a,c,k\}}$ as illustrated in Figure \ref{fig:grid2}.
In Figure \ref{fig:grid2}, observe that $g$ is a bottleneck with views $b,f,p$.
Furthermore, $g$ is label consistent for $\{a,c,k\}$ with a reference $b$ by assuming $\beta_{bg}>0$ (or $\beta_{bg}<0$).
Hence, one can recover $\mathbb{P}_{\beta,\gamma}\left(x_{\{b,f,g,p\}}|x_{\{a,c,k\}}\right)$ using $\mathtt{TensorDecomp}$ and obtain $\mathbb{P}_{\beta,\gamma}\left(x_{\{a,b,c,f,g,k,p\}}\right)$ using the following identity.
\begin{align*}
\mathbb{P}_{\beta,\gamma}\big(&x_{\{a,b,c,f,g,k,p\}}\big)=\mathbb{P}_{\beta,\gamma}\left(x_{\{b,f,g,p\}}|x_{\{a,c,k\}}\right)\mathbb{P}_{\beta,\gamma}\left(x_{\{a,c,k\}}\right).
\end{align*}
Similarly, one can recover $\mathbb{P}_{\beta,\gamma}\left(x_{\{a,f,k,\ell,p,q,r\}}\right)$, $\mathbb{P}_{\beta,\gamma}\left(x_{\{c,d,e,i,j,o,t\}}\right)$, $\mathbb{P}_{\beta,\gamma}\left(x_{\{e,j,n,o,r,s,t\}}\right)$.

In order to recover marginals including $x_h$ or $x_m$, $h$ and $m$ should be bottlenecks.
Conditioned on $x_{\{b,d,\ell,q\}}$, as illustrated in Figure \ref{fig:grid4}, $h$ is a bottleneck with views $c,p,r$, however, we do not have a marginal $\mathbb{P}_{\beta,\gamma}\left(x_{\{b,c,d,\ell,p,q,r\}}\right)$ currently.
Now, we recover the marginal $\mathbb{P}_{\beta,\gamma}\left(x_{\{b,c,d,\ell,p,q,r\}}\right)$.
Consider the distribution conditioned on $x_{\{p,r\}}$ as illustrated in Figure \ref{fig:grid3}.
In Figure \ref{fig:grid3}, observe that $q$ and $b,c,d$ are disconnected if $\ell$ is removed.
Furthermore, $\mathbb{P}_{\beta,\gamma}\left(x_{\{\ell,q\}}|x_{\{p,r\}}\right)$ and $\mathbb{P}_{\beta,\gamma}\left(x_{\{b,c,d,q\}}|x_{\{p,r\}}\right)$ are already observed.
Hence, using $\mathtt{LinearView}$ by setting $S\leftarrow\{b,c,d\}$, $i\leftarrow\ell$, $j\leftarrow q$ and conditioning $x_{\{p,r\}}$, one can obtain $\mathbb{P}_{\beta,\gamma}\left(x_{\{b,c,d,\ell,p,q,r\}}\right)$.
Now, $h$ is a bottleneck with views $c,p,r$ by conditioning $x_{\{b,d,\ell,q\}}$.
Using $\mathtt{TensorDecomp}$ one can obtain $\mathbb{P}_{\beta,\gamma}\left(x_{\{b,c,d,h,\ell,p,q,r\}}\right)$.
Using same procedure, one can also obtain $\mathbb{P}_{\beta,\gamma}\left(x_{\{a,b,c,g,m,q,r,s\}}\right)$.

Until now, we have recovered every pairwise marginals between visible variable and latent variable.
The remaining goal is to recover pairwise marginals between latent variables.
First, by setting $S\leftarrow\{e,j,o\}$, $i\leftarrow h$, $j\leftarrow c$ and conditioning $x_{\{b,d\}}$, one can recover $\mathbb{P}_{\beta,\gamma}\left(x_{\{b,c,d,e,h,j,o\}}\right)$ using $\mathtt{LinearView}$.
Consecutively, by setting $S\leftarrow\{h\}$, $i\leftarrow i$, $j\leftarrow j$ and conditioning $x_{\{e,o\}}$, one can recover $\mathbb{P}_{\beta,\gamma}\left(x_{\{e,h,i,j,o\}}\right)$ using $\mathtt{LinearView}$ which includes the pairwise marginals $\mathbb{P}_{\beta,\gamma}\left(x_{\{i,j\}}\right)$.
Other pairwise marginals between latent variables can be also recovered using the same procedure.
Since we end the sequence in 5 steps, this completes the proof of Lemma \ref{thm:grid}.

\section{Proof of Lemma \ref{lem:convRBM}}\label{sec:pflem:convRBM}

We first consider the distribution conditioned on $x_{\{c,e,f\}}$ as illustrated in Figure \ref{fig:crbm2}.
In Figure \ref{fig:crbm2}, observe that $m$ is a bottleneck with views $a,b,d$ with a reference $a$ by assuming $\beta_{am}>0$ (or $\beta_{am}<0$).
Hence, one can recover $\mathbb{P}_{\beta,\gamma}\left(x_{\{a,b,d,m\}}|x_{\{c,e,f\}}\right)$ using $\mathtt{TensorDecomp}$ and obtain $\mathbb{P}_{\beta,\gamma}\left(x_{\{a,b,c,d,e,f,m\}}\right)$ using the following identity.
\begin{align*}
\mathbb{P}_{\beta,\gamma}\big(&x_{\{a,b,c,d,e,f,m\}}\big)=\mathbb{P}_{\beta,\gamma}\left(x_{\{a,b,d,m\}}|x_{\{c,e,f\}}\right)\mathbb{P}_{\beta,\gamma}\left(x_{\{c,e,f\}}\right).
\end{align*}
Similarly, one can recover $\mathbb{P}_{\beta,\gamma}\left(x_{\{a,b,c,d,e,f,n\}}\right)$, $\mathbb{P}_{\beta,\gamma}\left(x_{\{g,h,i,j,k,\ell,q\}}\right)$, $\mathbb{P}_{\beta,\gamma}\left(x_{\{g,h,i,j,k,\ell,r\}}\right)$.

In order to recover marginals including $x_o$ or $x_p$, $o$ and $p$ should be bottlenecks.
Conditioned on $x_{\{h,m,q\}}$, $o$ is a bottleneck with views $d,e,g$, however we do not have a marginal $\mathbb{P}_{\beta,\gamma}\left(x_{\{d,e,g,h,m,q\}}\right)$ currently.
Now, we recover the marginal $\mathbb{P}_{\beta,\gamma}\left(x_{\{d,e,g,m,q\}}\right)$.
Since we observed $\mathbb{P}_{\beta,\gamma}\left(x_{\{a,b,d,e,g,h,j,k\}}\right)$ and $\mathbb{P}_{\beta,\gamma}\left(x_{\{a,b,d,e,m\}}\right)$, we can recover $\mathbb{P}_{\beta,\gamma}\left(x_{\{a,b,d,e,g,h,j,k,m\}}\right)$ using $\mathtt{DisjointView}$ by setting $S\leftarrow\{g,h,j,k\}$, $T\leftarrow\{m\}$ and $C\leftarrow\{a,b,d,e\}$.
Likewise, using $\mathtt{DisjointView}$, one can recover a marginal $\mathbb{P}_{\beta,\gamma}\left(x_{\{a,b,d,e,g,h,j,k,m,q\}}\right)$ as well.
Using the recovered marginal $\mathbb{P}_{\beta,\gamma}\left(x_{\{d,e,g,h,m,q\}}\right)$, conditioning $x_{\{h,m,q\}}$ and using $\mathtt{TensorDeomp}$, one can recover  $\mathbb{P}_{\beta,\gamma}\left(x_{\{d,e,g,h,m,o,q\}}\right)$.
Similarly, one can recover $\mathbb{P}_{\beta,\gamma}\left(x_{\{e,f,h,i,n,p,r\}}\right)$. 
Since we end the sequence in 4 steps, this completes the proof of Lemma \ref{thm:grid}.

\section{Proof of Lemma \ref{thm:regular}}\label{sec:pflem:regular}
The main idea of the proof is to show that every latent nodes of size $\le cN$ contains at least a single recoverable latent node using $\mathtt{TensorDecomp}$ where $N=|V|$.
We first state the following condition for a latent node $i$.
\begin{condition}\label{cond:regular}
For a latent node $i$, two of its neighbors $j,k$ are visible and a set of neighbors $S$ of $j,k$ are visible except for $i$, not containing $j,k$.
Also, there exists $\ell\in O\setminus S$ such that $i$ is a bottleneck with views $j,k,\ell$ in $G\setminus S$.
\end{condition}
In the above condition, $O$ denote the set of visible nodes.
One can easily observe that if any latent node satisfies the above condition, then it is recoverable by conditioning neighbors of $j,k$ and apply $\mathtt{TensorDecomp}$ with views $j,k$ and some other.

Now consider the following procedure.
First, duplicate for each $i\in V$ into $i_1,\dots,i_d$ where $i_n$ is visible/latent if $i$ is visible/latent.
Let $V^\prime$ be a such duplicated vertex set and $O^\prime\subset V^\prime$ be a set of visible nodes and $H^\prime=V^\prime\setminus O^\prime$ be a set of latent nodes.
The procedure starts with a graph on $V^\prime$ without edges.
\begin{itemize}
\item[1.] Choose latent nodes $i_1,\dots,i_d\in H^\prime$.
For each $n\in\{1,\dots,n\}$
if deg$(i_n)\ne 1$,
Choose a single neighbor $j_m$ of $i_n$ with probability
$$\mathbb{P}(j_m~\text{is chosen})=\frac{1-\text{deg}(j_m)}{\sum_{k_o\in V^\prime}(1-\text{deg}(k_o))}.$$
\item[2.]Similarly, for each neighbor $j_m\in O^\prime$ of $i_1,\dots,i_d$, for all $j_1,\dots,j_d$ satisfying deg$(j_o)=0$, add neighbors of $j_o$ as in step 1.
\item[3.] Check whether there exists an edge $(i_n,i_m)$ or a pair of edges $(i_n,j_m)$, $(i_{n^\prime},j_{m^\prime})$.
If such edge or a pair of edges exists, then the procedure restarts from the beginning.
\item[4.] Let $G$ be a graph such that contracting $\ell_1,\dots,\ell_d$ into $\ell$ for all $\ell\in V$. 
Check whether $i$ satisfies Condition \ref{cond:regular} with $j,k$ and $i$ is a bottleneck by conditioning neighbors of $j,k$.
\item[5.]If $G$ satisfies the condition in step 3, then the procedure succeeds. 
If not, repeat the procedure for the next latent node until every latent node decides its neighbor.
\item[6.] If every latent nodes decided its neighbor, the procedure fails.
\end{itemize}
The above procedure is constructing the fractional edges of random $d$-regular graph by contracting $\ell_1,\dots,\ell_d$ into $\ell$.
step 3 checks whether the procedure creates a loop or multiple edges.
One can notice that if any node satisfies Condition \ref{cond:regular} in step 3, then there exists a recoverable latent node.
Our primary goal is to bound the probability that the procedure fails, i.e., no latent node satisfies Condition \ref{cond:regular} under the fractional graph.

One can observe that if some visible node is chosen to be a neighbor of a latent node in the procedure but it is already a neighbor of other latent node, then it cannot help to satisfy Condition \ref{cond:regular}.
Also, at each iteration, choosing neighbor has an effect that reducing at most $2d$ nodes from whole nodes as at most $d^2$ edges are created.
Now, suppose there exist $\alpha n$ latent nodes where $\alpha<\frac1{2d(d-1)}$.
Using this fact, one can observe that 
the probability that a visible node connected to a latent node has $d-1$ visible neighbors is at least $p=(1-2d\alpha)^{d-1}$.
We also note that the probability that the procedure start over in step 3 is $O(1/N)$ at each iteration.
Therefore, one can conclude that

\begin{align}
&\mathbb{P}(\text{the procedure fails})\notag\\
&\le\prod_{i\in H}\Bigg[O(1)\sum_{n=0}^{d-\text{deg}(i)}\left(\frac{\alpha N}{(1-2d\alpha)N}\right)^{d-\text{deg}(i)-n}\left((1-p)^n+np(1-p)^{n-1}+O\left(\frac1N\right)\right)^{\mathbf{1}_{n\ge 2}} \Bigg]^{\mathbf{1}_{d-\text{deg}(i)\ge2}}\notag\\
&\le\prod_{i\in H}\Bigg[O(1)\sum_{n=0}^{d-\text{deg}(i)}\alpha^{d-\text{deg}(i)-n}\left(\alpha^{n-1}+O\left(\frac1N\right)\right)^{\mathbf{1}_{n\ge 2}}\Bigg]^{\mathbf{1}_{d-\text{deg}(i)\ge2}}\notag\\
&\le\prod_{i\in H}\Bigg[O(1)\alpha^{d-\text{deg}(i)-1}\Bigg]^{\mathbf{1}_{d-\text{deg}(i)\ge2}}\notag\\
&\le\left(O(1)\alpha^{d-1}\right)^{\alpha N/(d+1)}\left(O(1)\alpha^{d-2}\right)^{d\alpha N/(d+1)^2}\notag\\
&\le \left(O(1)\alpha\right)^{k\alpha N}\label{eq:regular}
\end{align}
for sufficiently small $\alpha$ (up to constant) 
where $O(1/N)$ in the bracelet represents the probability that non-existence of $\ell$ in Condition \ref{cond:regular} and the degree varies as the procedure iterates.
Also, $\mathbf{1}_S$ is an indicator function having a value $1$ if an event $S$ occurs, $0$ if not.
The second last inequality follows from the fact that we can choose at least $\alpha n/(d+1)$ latent nodes of degree $0$ at first, and then, we can choose at least $d\alpha n/(d+1)^2$ latent nodes of degree less than or equal to $1$.
$k$ in the last inequality is
$$k=\frac{2d^2-2d-1}{(d+1)^2}>1$$
for all $d\ge 5$.
One might concern that after the procedure succeeds, the extension of the procedure to the all vertices may start over with high probability so that the probability $\mathbb{P}(\text{no latent node satisfies Condition \ref{cond:regular}})$ becomes significantly larger than \eqref{eq:regular}.
However, we note that the restarting probability that extending the procedure to all vertices is $1-\exp\left(\frac{1-d^2}{4}\right)$ a.a.s., i.e., constant, (see \cite{wormald1999models}) and therefore 
$$\mathbb{P}(\text{no latent node satisfies Condition \ref{cond:regular}})\le\exp\left(\frac{d^2-1}{4}\right)\mathbb{P}(\text{the procedure fails})=\left(O(1)\alpha\right)^{k\alpha N}$$
for $O(1)\alpha<1$ in the above equation.
Now, we consider all $1\le\alpha N\le cN$ and all choices of sets of latent node to apply the union bound as below.
The explicit choice of $c$ will be presented later.
\begin{align*}
&\mathbb{P}(\text{no latent node satisfies Condition \ref{cond:regular} for all choices of a set of latent node with $1\le\alpha N\le cN$})\\
&\qquad=\sum_{1\le\alpha N\le cN}\binom{N}{\alpha N}\mathbb{P}(\text{no latent node satisfies Condition \ref{cond:regular}})\\
&\qquad\le\sum_{1\le\alpha N\le cN}\binom{N}{\alpha N}\left(O(1)\alpha\right)^{k\alpha N}\\
&\qquad\le\sum_{1\le\alpha n\le cn}O(1)\sqrt{\frac{1}{\alpha(1-\alpha)n}}\alpha^{-\alpha n}(1-\alpha)^{-(1-\alpha)n}\left(O(1)\alpha\right)^{k\alpha n}\\
&\qquad\le\sum_{1\le\alpha n\le cn}O(1)\sqrt{\frac{1}{\alpha(1-\alpha)n}}\exp\Big[\Big((k-1)\alpha\log\alpha+\alpha O(1)-(1-\alpha)\log(1-\alpha)\Big)n\Big]\\
&\qquad=o(1)
\end{align*}
where the first inequality is from Stirling's formula and we choose $c$ to satisfy that $(k-1)c\log c+c\log O(1)-(1-c)\log(1-c)<0$ to obtain the last equality.
Such $c$ always exists as
\begin{align*}
(k-1)c\log c+c\log O(1)-(1-c)\log(1-c)=c((k-1)\log c+O(1)+1-O(c))<0
\end{align*}
for a sufficiently small $c$.

Now, we know that at each iteration of the sequential learning framework, there exists at least one bottleneck latent node which can be recovered without labeling issue (forcing labels).
Furthermore, using $\mathtt{LinearView}$ and conditioning, one can also treat recovered latent nodes as visible nodes while the marginals including latent nodes always containing the conditioned variables, i.e., the order of marginals reduces in some sense as recovered marginals has fixed order while a part of order is the constant number (at most $d-1$) of conditioned variables.
Using this fact, one can conclude that the sequential learning framework recovers every pairwise marginals in $2d|H|$ iterations. where $2d$ follows from that the upperbound of calls of $\mathtt{LinearView}$ for recovering a single latent node is $2d-2$ and at most two bottleneck calls are required.
This completes the proof of Theorem \ref{thm:regular}.

%% file: main_arxiv.bbl
\begin{thebibliography}{10}

\bibitem{Anandkumar2012}
Animashree Anandkumar, Dean~P Foster, Daniel~J Hsu, Sham~M Kakade, and Yi~kai
  Liu.
\newblock A spectral algorithm for latent dirichlet allocation.
\newblock In {\em Advances in Neural Information Processing Systems}, pages
  917--925, 2012.

\bibitem{anandkumar2014tensor}
Animashree Anandkumar, Rong Ge, Daniel~J Hsu, Sham~M Kakade, and Matus
  Telgarsky.
\newblock Tensor decompositions for learning latent variable models.
\newblock {\em Journal of Machine Learning Research}, 15(1):2773--2832, 2014.

\bibitem{anandkumar2012method}
Animashree Anandkumar, Daniel~J Hsu, and Sham~M Kakade.
\newblock A method of moments for mixture models and hidden markov models.
\newblock In {\em Conference on Learning Theory}, 2012.

\bibitem{bach2002kernel}
Francis~R Bach and Michael~I Jordan.
\newblock Kernel independent component analysis.
\newblock {\em Journal of machine learning research}, 3(Jul):1--48, 2002.

\bibitem{Balle2012}
Borja Balle and Mehryar Mohri.
\newblock Spectral learning of general weighted automata via constrained matrix
  completion.
\newblock In {\em Advances in Neural Information Processing Systems}, pages
  2159--2167, 2012.

\bibitem{Chaganty2013}
Arun~T. Chaganty and Percy Liang.
\newblock Spectral experts for estimating mixtures of linear regressions.
\newblock In {\em International Conference on Machine Learning}, pages
  1040--1048, 2013.

\bibitem{chaganty2014estimating}
Arun~T. Chaganty and Percy Liang.
\newblock Estimating latent-variable graphical models using moments and
  likelihoods.
\newblock In {\em International Conference on Machine Learning}, pages
  1872--1880, 2014.

\bibitem{comon2010handbook}
Pierre Comon and Christian Jutten.
\newblock {\em Handbook of Blind Source Separation: Independent component
  analysis and applications}.
\newblock Academic press, 2010.

\bibitem{dempster1977maximum}
Arthur~P Dempster, Nan~M Laird, and Donald~B Rubin.
\newblock Maximum likelihood from incomplete data via the em algorithm.
\newblock {\em Journal of the royal statistical society. Series B
  (methodological)}, pages 1--38, 1977.

\bibitem{fayyad1996data}
Usama Fayyad, Gregory Piatetsky-Shapiro, and Padhraic Smyth.
\newblock From data mining to knowledge discovery in databases.
\newblock {\em AI magazine}, 17(3):37, 1996.

\bibitem{freeman2000learning}
William~T Freeman, Egon~C Pasztor, and Owen~T Carmichael.
\newblock Learning low-level vision.
\newblock {\em International journal of computer vision}, 40(1):25--47, 2000.

\bibitem{gallager1962low}
Robert Gallager.
\newblock Low-density parity-check codes.
\newblock {\em IRE Transactions on information theory}, 8(1):21--28, 1962.

\bibitem{Halpern2013}
Yoni Halpern and David Sontag.
\newblock Unsupervised learning of noisy-or bayesian networks.
\newblock In {\em Uncertainty in Artificial Intelligence}, pages 272--281,
  2013.

\bibitem{hinton2002training}
Geoffrey~E Hinton.
\newblock Training products of experts by minimizing contrastive divergence.
\newblock {\em Neural computation}, 14(8):1771--1800, 2002.

\bibitem{Hsu2013}
Daniel Hsu and Sham~M. Kakade.
\newblock Learning mixtures of spherical {G}aussians: Moment methods and
  spectral decompositions.
\newblock In {\em Innovations in Theoretical Computer Science}, 2013.

\bibitem{hsu2012spectral}
Daniel Hsu, Sham~M Kakade, and Tong Zhang.
\newblock A spectral algorithm for learning hidden markov models.
\newblock {\em Journal of Computer and System Sciences}, 78(5):1460--1480,
  2012.

\bibitem{Hyvarinen_2000}
A.~Hyv\"{a}rinen and E.~Oja.
\newblock Independent component analysis: Algorithms and applications.
\newblock {\em Neural Networks}, 13(4-5):411--430, 2000.

\bibitem{jordan1998learning}
Michael~Irwin Jordan.
\newblock {\em Learning in graphical models}, volume~89.
\newblock Springer Science \& Business Media, 1998.

\bibitem{kschischang1998iterative}
Frank~R. Kschischang and Brendan~J. Frey.
\newblock Iterative decoding of compound codes by probability propagation in
  graphical models.
\newblock {\em IEEE Journal on Selected Areas in Communications},
  16(2):219--230, 1998.

\bibitem{lee2009convolutional}
Honglak Lee, Roger Grosse, Rajesh Ranganath, and Andrew~Y Ng.
\newblock Convolutional deep belief networks for scalable unsupervised learning
  of hierarchical representations.
\newblock In {\em International Conference on Machine Learning}, pages
  609--616, 2009.

\bibitem{mossel2005learning}
Elchanan Mossel and S{\'e}bastien Roch.
\newblock Learning nonsingular phylogenies and hidden markov models.
\newblock In {\em Proceedings of the thirty-seventh annual ACM symposium on
  Theory of computing}, pages 366--375. ACM, 2005.

\bibitem{Parikh2012}
Ankur~P. Parikh, Le~Song, Mariya Ishteva, Gabi Teodoru, and Eric~P. Xing.
\newblock A spectral algorithm for latent junction trees.
\newblock In {\em Uncertainty in Artificial Intelligence}, pages 675--684,
  2012.

\bibitem{Parikh2011}
Ankur~P. Parikh, Le~Song, and Eric~P. Xing.
\newblock A spectral algorithm for latent tree graphical models.
\newblock In {\em International Conference on Machine Learning}, pages
  1065--1072, 2011.

\bibitem{parisi1988statistical}
Giorgio Parisi and Ramamurti Shankar.
\newblock Statistical field theory, 1988.

\bibitem{podosinnikova2015rethinking}
Anastasia Podosinnikova, Francis Bach, and Simon Lacoste-Julien.
\newblock Rethinking lda: moment matching for discrete ica.
\newblock In {\em Advances in Neural Information Processing Systems}, pages
  514--522, 2015.

\bibitem{redner1984mixture}
Richard~A Redner and Homer~F Walker.
\newblock Mixture densities, maximum likelihood and the em algorithm.
\newblock {\em SIAM review}, 26(2):195--239, 1984.

\bibitem{salakhutdinov2009deep}
Ruslan Salakhutdinov and Geoffrey~E Hinton.
\newblock Deep boltzmann machines.
\newblock In {\em International Conference on Artificial Intelligence and
  Statistics}, volume~1, page~3, 2009.

\bibitem{siddiqi2010reduced}
Sajid~M Siddiqi, Byron Boots, and Geoffrey~J Gordon.
\newblock Reduced-rank hidden markov models.
\newblock In {\em International Conference on Artificial Intelligence and
  Statistics}, pages 741--748, 2010.

\bibitem{song2014nonparametric}
Le~Song, Animashree Anandkumar, Bo~Dai, and Bo~Xie.
\newblock Nonparametric estimation of multi-view latent variable models.
\newblock In {\em International Conference on Machine Learning}, pages
  640--648, 2014.

\bibitem{song2010hilbert}
Le~Song, Byron Boots, Sajid~M Siddiqi, Geoffrey~J Gordon, and Alex~J Smola.
\newblock Hilbert space embeddings of hidden markov models.
\newblock In {\em International Conference on Machine Learning}, pages
  991--998, 2010.

\bibitem{song2011kernel}
Le~Song, Eric~P Xing, and Ankur~P Parikh.
\newblock Kernel embeddings of latent tree graphical models.
\newblock In {\em Advances in Neural Information Processing Systems}, pages
  2708--2716, 2011.

\bibitem{wainwright2008graphical}
Martin~J Wainwright, Michael~I Jordan, et~al.
\newblock Graphical models, exponential families, and variational inference.
\newblock {\em Foundations and Trends{\textregistered} in Machine Learning},
  1(1--2):1--305, 2008.

\bibitem{wormald1999models}
Nicholas~C Wormald.
\newblock Models of random regular graphs.
\newblock {\em London Mathematical Society Lecture Note Series}, pages
  239--298, 1999.

\bibitem{zhang2015spectral}
Chicheng Zhang, Jimin Song, Kamalika Chaudhuri, and Kevin Chen.
\newblock Spectral learning of large structured hmms for comparative
  epigenomics.
\newblock In {\em Advances in Neural Information Processing Systems}, pages
  469--477, 2015.

\bibitem{zou2013contrastive}
James~Y Zou, Daniel~J Hsu, David~C Parkes, and Ryan~P Adams.
\newblock Contrastive learning using spectral methods.
\newblock In {\em Advances in Neural Information Processing Systems}, pages
  2238--2246, 2013.

\end{thebibliography}
